\documentclass[conference]{IEEEtran}
\IEEEoverridecommandlockouts
\usepackage{cite}
\usepackage{amsmath,amssymb,amsfonts}
\usepackage{algorithmic}
\usepackage{graphicx}
\usepackage{textcomp}
\usepackage{xcolor}
\usepackage{subcaption}
\usepackage{stfloats}
\usepackage{balance}

\def\BibTeX{{\rm B\kern-.05em{\sc i\kern-.025em b}\kern-.08em
    T\kern-.1667em\lower.7ex\hbox{E}\kern-.125emX}}
\begin{document}

\title{
How saccadic vision might help with the interpretability of deep networks\\
\thanks{The study was supported by the Ministry of Education and Science of Russia (Project No. 14.Y26.31.0022).}
}
\makeatletter
\newcommand{\linebreakand}{%
  \end{@IEEEauthorhalign}
  \hfill\mbox{}\par
  \mbox{}\hfill\begin{@IEEEauthorhalign}
}
\makeatother

\author{\IEEEauthorblockN{1\textsuperscript{st} Iana Sereda}
\IEEEauthorblockA{\textit{Research and Education Center }\\
\textit{Mathematics of Future Technologies}\\
Nizhny Novgorod, Russia \\
sereda@itmm.unn.ru}

\and
\IEEEauthorblockN{2\textsuperscript{nd} Grigory Osipov}
\IEEEauthorblockA{\textit{Research and Education Center }\\
\textit{Mathematics of Future Technologies}\\
Nizhny Novgorod, Russia \\
osipov@vmk.unn.ru}
}

\maketitle

\begin{abstract}
We describe how 
some problems (interpretability, lack of object-orientedness) of modern deep networks potentially could be solved by adapting a biologically plausible saccadic mechanism of perception. A sketch of such a saccadic vision model is proposed. Proof of concept experimental results are provided to support the proposed approach.
\end{abstract}


\section{Introduction}
The topic of interpretability of data representation arising in an intelligent agent is interesting for many reasons. For example, medical applications require  decision-making to be transparent for a human because the cost of a mistake is high. Again, the higher the interpretability of the representation for a human, the more it is similar to how a human thinks, and, therefore, the closer we are to solving the problem of modeling intelligence in general. 

The converse is also true: data on the mechanisms of the biological brain can be useful in the sense of introducing their principles into artificial intelligence models. Thus, some features of the structure of the visual cortex inspired the development of convolutional networks\cite{lindsay2020convolutional}. 

\begin{figure}[!ht]
\centering
\includegraphics[scale=0.5]{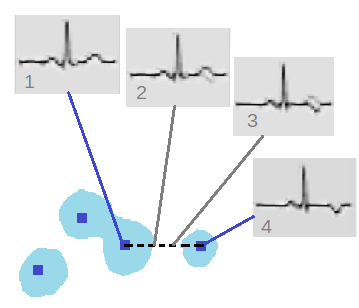}
\caption{\small{A schematic example of the presence of non-interpretable zones in the code space of a trained convolutional neural network. Blue dots are projections of real electrocardiogram (ECG) fragments into the code space of the model. The blue area around them is the area of interpretable codes, the white area is of not interpretable codes. ECG pictures are generated by a 2d convolutional generative model from points in code space located on a dashed line segment. The input data for the convolutional autoencoder were images of the cardiac cycle, considered as a two-dimensional low-resolution black and white picture. It can be seen that pictures 2 and 3 show impossible heart cycles.}}
\label{zones_interpretability}
\end{figure}

Deep convolutional networks are often used today in applied problems as one of the basic components of learning systems. On some tasks, for example, the task of modeling faces, it is possible to achieve representations with good interpretability\cite{karras2019style}. However, this success does not translate to all types of data equally.  Problems begin in those tasks where generalization is required for a small amount of data; where datasets are poorly balanced; where active learning is required and catastrophic forgetting must be avoided\cite{buda2018systematic}\cite{kemker2018measuring}; A separate type of difficulty is the selection of architecture hyperparameters\cite{he2021automl}.
Finally, it is known that the representation of deep convolutional networks is not object-oriented(details below), while in human representation proto-objects appear already at the early stages of processing of visual information\cite{von2015figure}. 
 This paper will examine how the use of the principles of saccadic vision might be helpful in fixing some of the above problems.

\section{Related Work}
One of the most human-readable representations of a visual scene is the semantic scene graph: if it is present, the task of generating the text describing the scene is trivial\cite{krishna2017visual}. The nodes of such a graph are usually nouns that name objects on the stage. The node can be assigned its coordinates on the image.
There are approaches to generate a scene graph using deep learning techniques. The first step of the algorithm is usually to select areas of the image in which the presence of objects is likely \cite{yang2018graph} \cite{xu2017scene}\cite{raboh2020differentiable}. Such regions are searched using a separate convolutional network, for example, Faster R-CNN\cite{ren2016faster}, which returns the coordinates of the region of interest and the corresponding feature tensor.
However, with this approach, graph representations arise not at every level of abstraction, but only at the highest, i.e. at the level of whole objects, but not at the level of their parts. Before the appearance of this graph, intermediate scene representations are activation tensors that arise in a neural network (or a group of neural networks), often convolutional.

Convolutional networks associate an image area with a feature tensor in which the concept of coordinates does not appear explicitly. When an attempt is made to obtain the coordinates of an object from the convolutional network in an explicit form, overfitting inevitably occurs, which is shown in \cite{liu2018intriguing}. 

Meanwhile, the concept of objects is inseparable from the concept of their coordinates. When modeling the world by physics, distances appear in the model of the world in an explicit form. The fact that distances do not explicitly occur in standard convolutional networks representations means that it is not guaranteed for representation to be object-oriented.

Distances are more explicitly generated when deep learning models are equipped with attention mechanisms. For example, in the model of recurrent attention a sequence of views is formed, whose coordinates can be obtained explicitly if desired\cite{mnih2014recurrent}. However, training models with attention, both old and newer\cite{vaswani2017attention}, requires a lot of data, since work is still being done with large networks. Thus, the task of creating a deep agent, whose data representation is object-oriented at all levels of abstraction and at the same time allows learning from several samples, has not yet been solved.

\subsection{Eye movements}
The retina of the eye has an uneven distribution of receptors, and therefore at any given time a person sees clearly only a relatively small part of the picture. Thus, in order to see the whole scene, a person makes many quick eye movements, saccades. At least in part, these movements are involuntary. Human saccades occur several times per second\cite{fabius2020low}. 
The generated sequence of saccades depends on the image being viewed and on the problem solved by the subject\cite{yarbus2013eye}.
The length of the focus movement during saccades is different, sometimes microsaccades are distinguished\cite{tian2018dynamics}.

From the above it follows that recognition of a scene by a human is a sequence of controls (of refocusing of the eyes), each of which leads to new sensory data. The brain that recognizes the scene does not see the new scene immediately, but learns it through the controls and their results. This is different from how the input image is represented when working with deep learning networks: the image is often thought of as a point in multidimensional vector space.

\subsection{Active exploration of a situation}
In some works, for example\cite{friston2012perceptions}, it is proposed to consider saccadic vision as a process of sequential generation of hypotheses by the brain and their immediate experimental verification. In this formulation, the saccade is the most informative experiment aimed at reducing the uncertainty about the structure of the situation. 

An active exploration of the situation by the biological brain occurs not only during the movement of the gaze, but also during the movement of the body, for example, when a rat is exploring a new environment. The representation of the environment in this case is built on the basis of place cells and grid cells\cite{moser2008place}. 
The module, consisting of grid cells, renews its activity as the animal moves around the environment and "marks on the map" where the animal is now.  There are grounds\cite{long2018novel} for the hypothesis that the place cells and grid cells are a particular version of a more general mechanism for modeling of the world by the mammalian brain\cite{friston2012perceptions}. If this is the case, then not only the embeddings of sensory data, but also the controls that led to this data, should be included in the representation of data at all levels of abstraction in the deep agent.

\section{Electrocardiograms}
Our numerical modeling was carried out on human electrocardiograms (ECG) as training data. Such data was chosen because in has a lower dimensionality than images of real scenes, but at the same time ECG still represents a scene with objects of different types placed on it. The objects here are not people, animals, trees, as it happens in photographs, but components of the human cardiac cycle: various waves and wave complexes. The appearance of these "objects" varies under the influence of many factors\cite{murata1992effects}: the patient's diseases, sliding of the electrodes on the skin, breathing, muscle contraction under the electrodes, age, gender, and even drinking cold water before the examination.

Deep networks have shown their applicability to many tasks that arise when working with ECG: the task of delineation\cite{malali2020supervised}, diagnosing certain pathological conditions\cite{acharya2017application}, compression \cite{yildirim2018efficient}. However, studies of the representation of ECG in convolutional networks show problems when interpolating between points in the code space\cite{sereda2020problems}.

\section{The proposed approach}
An interesting point about grid cells is the lack of influence of external sensory data (for example, visual data) on the activity of cells of that type. Grid cells work only on information about the animal's own movements\cite{hafting2005microstructure}. 
Thus the mechanism of grid cells can be considered as the assignment of equivalence classes on the set of available controls. I.e. being at A, the animal can reach location B, performing different sequences of movements. The fact that they all lead to the same outcome, unifies them into an outcome equivalence class. 

This data prompts us to consider the assumption that the representation of a data in the brain is a simulation of sensory results of controls. Both types of data - data on controls and on their sensory results - must be present in the representation of the intectual agent in order to increase its interpretability. In other words problem of dimensionality reduction of sensory data probably should not be considered in isolation from the controls at which such sensory data are observed. In the case of vision, controls are saccades, in the case of navigation in space of body in space controls are complicated series of body movements. 
\subsection{Displacement of an object within the receptive field of the network}
Many types of data people perceive as a scene of objects that interact, forming a hierarchical structure. Objects inside the scene can change their relative positions from situation to situation. The response of the deep convolutional networks to the shift (of the object inside the receptive field) has negative characteristics. The conceptual problem is depicted in Fig. \ref{shift_probelm}.
\begin{figure}[!ht]
\centering
\includegraphics[scale=0.25]{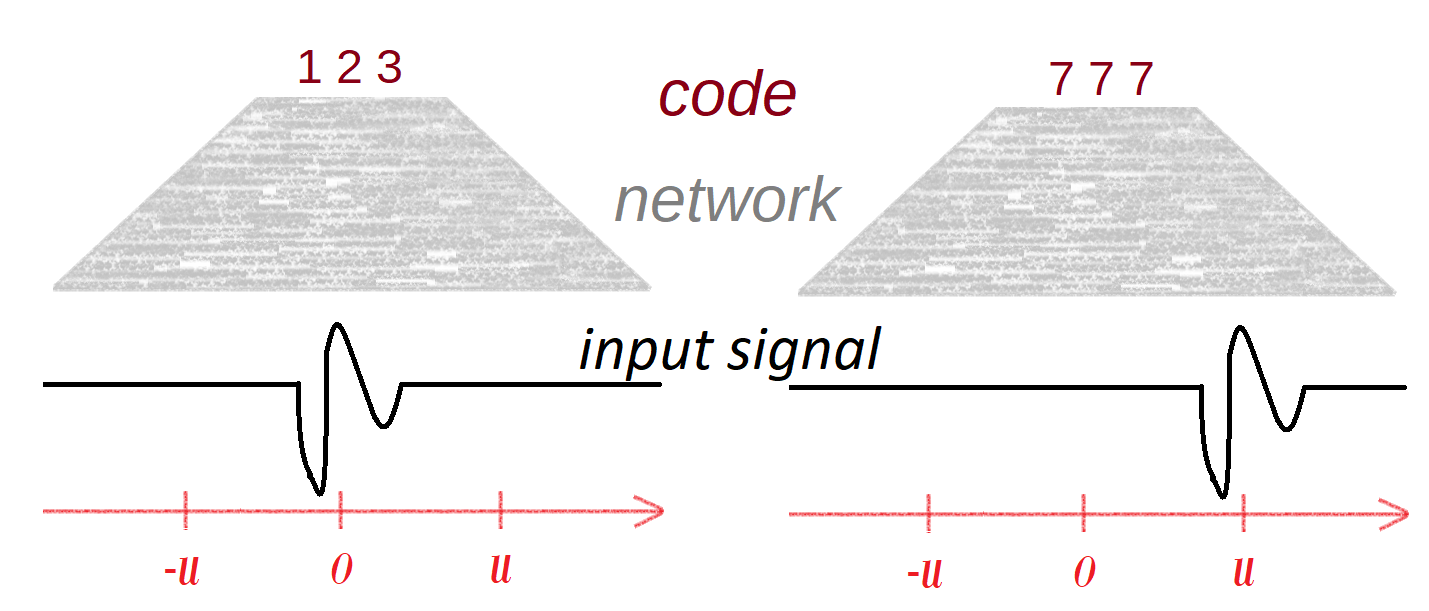}
\caption{\small{An illustration of the appearance of uninterpretability in the deep layer. In the first case (left), the specific pattern is in the center of the receptive field. In the second case, it is shifted (right). Despite the fact that the object remains the same, in the general case, the codes in the deep layers may be different. This is not necessarily always the case, but in standard deep convolutional network there is no guarantee against this behavior. Thus, morphology information becomes inherently entangled with displacement information. }}
\label{shift_probelm}
\end{figure}
In \cite{zhang2019making} it is shown that its reasons, at least in part, lie in the method of dimensionality reduction. I.e. in the pooling operations that are a standard part of convolutional architectures.

As a conceptual way to solve the above problem, we propose the separation of information about "what is observed" and "under what control" into two types of variables inside the representation of an intelligent agent. 
Fig. \ref{toy_exmple}. shows how, using the example of the situation from Fig. \ref{shift_probelm}, it is possible to correct problematic response to a shift. 
\begin{figure}[!ht]
\centering
\includegraphics[scale=0.25]{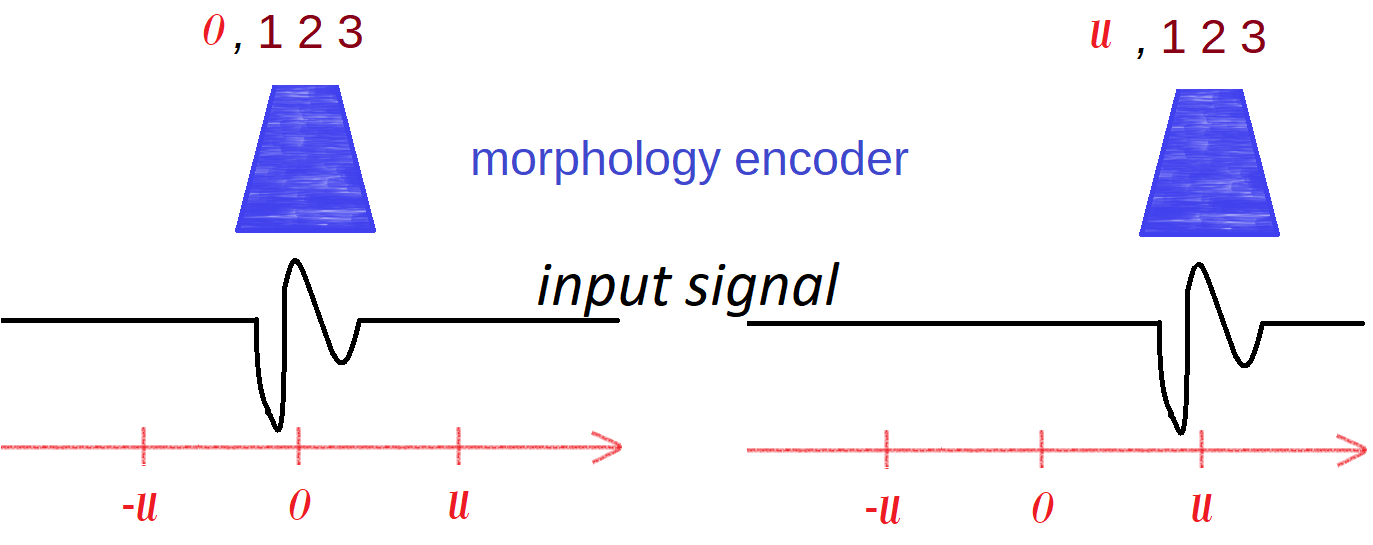}
\caption{\small{Representation in which information about the attributes of an object and about its location in scene coordinates is divided into two different variables.}}
\label{toy_exmple}
\end{figure}

In the above figure, there is an instance of a certain class of objects - a splash of a characteristic shape. There is also a "blue device" - a specialist in this type of objects (it can be a small neural network or another tool for dimensionality reduction). The device gives an informative description of the object seen, "123". The description it gives is informative if and only if the object being described is in the center of its receptive field. Therefore, one needs to apply the correct control to move the blue device so that it is in the right place. The pair "control" — "sensory result" here is analogous to the visual concepts: saccade — fixation.

\subsection{Uncertainty of controls}
Objects in the real world are not always located at fixed distances from each other. Consider photographs of people. Sometimes there may be a tree next to the person in the photo. If our gaze is looking at the object named "the left eye of a person", it is impossible to predict where to move the gaze so that a tree falls into its center. After all, the tree may not be present in this photo. Even if it is present, it can be anywhere.

But let's ask another question. Where do we need to shift our gaze so that to see the nose of this man? This question \textit{can} be answered in contrast to the question about a tree. The answer will not be deterministic, it will be approximate. 

Thus, the "saccadic" face model does not have to generate an accurate prediction of where to move the gaze in order for it to capture the next landmark. Usually accurate coordinate prediction is not possible even in the case of such a rigid structure as a face. However, the model could suggest not the exact coordinate, but the region the next one is in. If this region is small enough, each point can be checked for the presence of the center of the next predicted next landmark using the brute-force search. According to our hypothesis this could correspond to mammalian microsaccades.

\subsection{Characteristic controls}
In the above example one could roughly predict where the nose is. But it is almost impossible to predict what would be seen outside of the facial area. There may be a tree, an ocean, a wall, etc. 

That means that for each landmark there are, roughly, two types of controls: with a predictable result and with an unpredictable one. The controls with predictable result we will call\textit{ characteristic }controls for that landmark.

\subsection{Uncertainty of predicted result of a control}
Control results can be predicted with varying degrees of accuracy even in the case of characteristic controls. For example, seeing a nose at a given moment, we can predict where the eye will be on the face in question, but we cannot predict its color, its iris pattern, its shape, the length of the eyelashes. That means that the accuracy of the prediction cannot be measured, say, as the pixel-by-pixel difference between the predicted data and actually discovered during the saccade. 

It is reasonable to measure the predictability of control results using the concept of entropy. For a given landmark $A$ and control $ u(A)$, one can ask the question: how much is the entropy of its results lower than the entropy of the results of an arbitrary control $v(A)$?

\subsection{Recurrent definition}\label{sec::recurr}
If $H(u(A))<< H(v(A))$, than $u(A)$ is a characteristic for $A$. Here A is an indicator function that has the meaning of a detector of a given type of landmarks. The set $u(A)$ is a set of $n$-dimensional vectors, each of which is obtained as a sensory result of control $u$, performed "from A". I.e. relative to the coordinate system specified by the triggering of the indicator function A.

If a vector random variable (of dimensionality $n$) has a low entropy of the empirically observed sample, then it is reasonable to make a hypothesis that the process of its sampling can be approximated as the generation of points from the vicinity of a low-dimensional structure embedded in $R^n$. In the simplest case, this low-dimensional structure is a point from n-dimensional space. In a more complex way, this may be a piecewise linear approximation of a manifold by a small ReLU network.  

Hence, in order to find out whether the control $u$ is characteristic for indicator $A$, we can solve the dimensionality reduction problem, trying to minimize the accompanying loss of information. With luck, we get an approximation of this low-dimensional structure around which the results for $u(A)$ are centered. For simplicity, but without loss of generality, we assume that the results $u(A)$ are concentrated around the point $b \in R^n$.

Then the execution of the saccade $u$ from the indicator $A$ is equivalent to testing the hypothesis of the following form "if from object $A$ we move our gaze to point  $u(A)$, then what we see there will be a point located close to $b$. Here, roughly, there are two options: the observed point $b'$ will be close to $b$ or not close. The presence of two options gives a direct reason to introduce the indicator function $B$:
$$
B(b')=
\begin{cases}
0 ,    dist(b,b') > threshold\\
1 ,    dist(b,b') \leq  threshold
\end{cases}
$$
We have described how to obtain new indicator from existing indicator through finding of characteristic control. This gives the rule of recursive formation of the proposed structure of "control - results of control"  memory of an intelligent agent. For a number of ECGs, the statement $AuB$ will be true after some correction for $u$ (i.e., after a series of microsaccades in the vicinity of $u$). For a part of the ECGs,  the statement will turn out to be incorrect. For example, instead of it, $Au!B$ will work. Binary coding success/failure is a property of indicators, and thus we got a composite indicator $AuB$, which takes on the values true or false. Now we can "ask the world" for all situations in which this indicator is triggered, and then look for characteristic controls for it ...

\subsection{Initial indicator}
To define a recurrent process, in addition to the stepwise rule of obtaining one indicator from another, an initial indicator is required. One can think of different heuristic ways to get it. A particular option is to get it from the teacher as an outer pointer of attention. For example, in textbooks on ECG interpretation, arrows often show where to look at a given picture of ECG in order to learn a new morphological feature.

\section{ECG experiments}

ECG is a type of data to which deep neural networks are successfully applied in the following sense: with a sufficiently large training dataset for some types of pathologies of the cardiovascular system, it is possible to obtain good quality indicators for some accepted formal quality metrics. It is shown that a convolutional network can effectively reduce the dimensionality of the ECG signal, but at the same time it has problems with representation in hidden layers\cite{sereda2020problems}. The representation of data is not sufficiently interpretable, and the functionalist doctors are not replaced yet. The successes of convolutional networks on formal metrics and the failure of interpretability on this data type prompts the study of new ways of constructing deep representations that would take this into account. So this data type was chosen by us for conducting minimal proof-of-concept experiments.

The experiments were carried out on the LUDB dataset\cite{kalyakulina2020ludb}. It contains approximately 200 ECGs with delineation of the cardiac cycle structure. In particular, the highest points of the R-peaks are marked. A small subset of healthy patients was selected.

\subsection{Indicator A}
First, we experimentally demonstrated that the precise centering of an object in the center of the receptive field when decreasing the dimensionality is very important. The R-peaks were the test object in this experiment.

R-peaks of healthy people are similar, but not exactly. Even in the same ECG the R-peaks may slightly change from cycle to cycle. However, due to their similarity the dimensionality reduction algorithms should work effectively if small ECG fragments containing the R-peak are fed to them as a data set.

Let's collect the dataset as follows: we select short signal fragments containing R-peaks from the ECGs of healthy people (from the first lead). Let's choose the fragments short enough so that little gets there, except for the R-peak and its small vicinity (for the LUDB dataset that signal duration is about 40 numbers). Thus, we got a small dataset of 40-dimensional points, each of which contains an R-peak, and the center of the peak is located in the 20-th component of the vector, that is, in the center of the clipped fragment. If the saccade (shift of the gaze) of the agent each time brings his gaze exactly to the center of the  R-peak, then the dataset for the training of an indicator specialized in R-peak morphologies would be just that.

We compared the fragments selected in this way with the fragments of the same duration (40), but obtained not by shifts of the gaze to the R-peaks, but by shifts of the gaze to random places of the ECGs.

 The entropy of this second set should be higher, because there is more uniformity among R-peaks than among random fragments. This can be seen when decreasing the dimensionality of the geminate dataset, as shown in Fig. \ref{control_dispersion}, left: R-peaks form a pronounced cluster when the dimensionality decreases from 40 to 2.
 
 Since this cluster is well expressed, it is easy to create an indicator  $A$ on its basis in the terminology of the discussed framework: the indicator $A$ will be considered triggered if the signal fragment received at its input falls into a circle approximating the blue cluster.
 
 However, the situation with dimensionality reduction quickly changes if, when cutting fragments with R-peaks, the center of the R-peak is not located\textit{ exactly} in the same place of the fragment, but \textit{approximately} in the same place.

To demonstrate that we added gaussian noise of varying strength to the R-peaks coordinates given by the human doctor and studied how this affects the quality of dimensionality reduction. Figure \ref{control_dispersion} shows that with increasing noise in control $u$, the possibility of dimensionality reduction effectively degrades. Even with modern methods of dimensionality reduction, linear separability (blue from red in fig.\ref{control_dispersion}) is becoming increasingly difficult to achieve. 
\begin{figure*}[ht]
\centering
\includegraphics[scale=0.19]{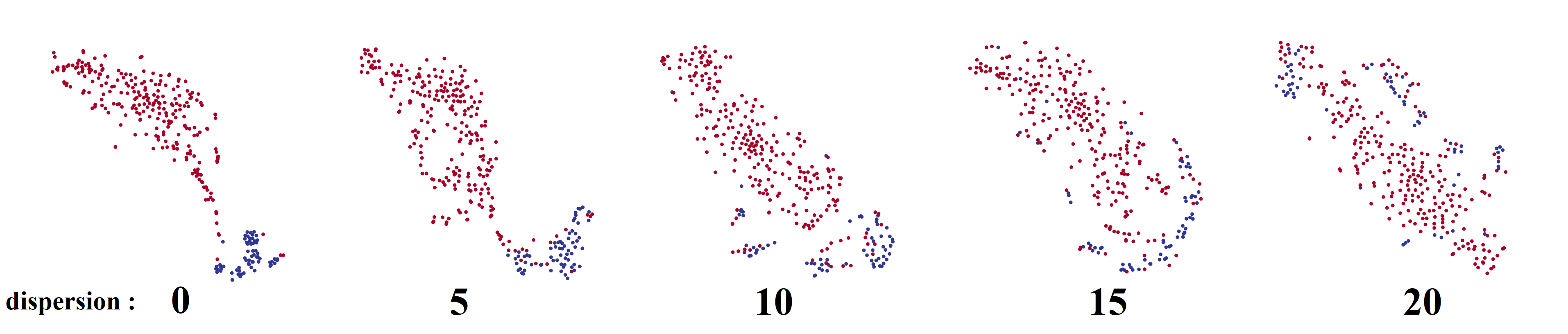}
\caption{\small{The result of clustering of 40-point ECG segments, lead i. All dots are  projections of these segments after dimensionality reduction by tSNE. Red dots corresponds to the "general population", i.e. to ECG segments cutted out without any system. The blue dots correspond to the purposefully cutted ECG segments: always in the vicinity of the R-wave in 7 healthy patients. The clustering result varies depending on how the "blue" segments were cutted out. In the far left picture the cutting occurs so that the center of the selected segment always coincides with actual center of the R-peak, which is indicated by the human expert. In the rest of the pictures, the cutting takes place in such a way that the center of the cutted segment is drawn from the normal distribution with the expectation at the actual center of the R-peak, but with non-zero variance. An increase in dispersion rapidly leads to degradation of the pattern of separable clusters. }}
\label{control_dispersion}
\end{figure*}

\begin{figure*}[b]
\centering
\includegraphics[scale=0.175]{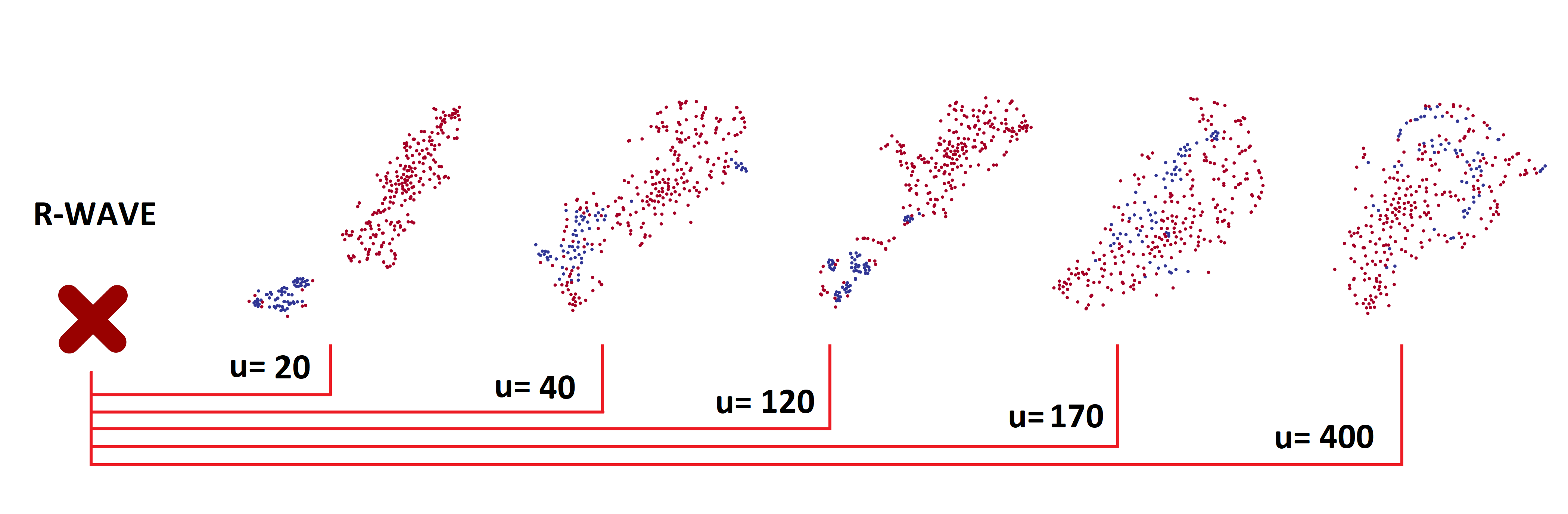}
\caption{\small{Results of clustering ECG segments depending on control. Control here means the offset length to the left of the R-peak. The blue dots in each figure correspond to ECG segments, the center of which is calculated as the center of the R-peak + offset $u$. The offset length at which the next set of "blue" segments was selected is shown above the arrows. Red color, as before, marks the ECG segments, selected completely at random from random ECGs, lead i. The length of the section (for both blue and red) is 40. This figure illustrates the fact that with most controls the cluster structure is not clearly expressed. However, under some controls it is clearly expressed and even is close to the linear separability of the red and blue clusters. In particular, clusters are well distinguishable at small $u$ (see $u= 20$) and  in the region of 120 (see $u=120$). Also clustering at large shifts is different from  clustering at the small shifts. At large shifts (see $u=170, 400$)  red and blue dots are distributed in a more similar way than in clustering at small shifts (see $u=20, 40, 120$), where  blue dots follow clearly different distribution compared to the red ones. }}
\label{different_controls}
\end{figure*}

\subsection{Characteristic control for indicator A}
Next, we conducted an experiment to find characteristic controls $u$ for this initial indicator $A$. We investigated the effectiveness of dimensionality reduction for the results of different controls for the above mentioned R-peak indicator. The qualitative result is shown in Fig.\ref{different_controls}. In this case, controls are gaze shifts on different distances from the center of the coordinate system (which is specified by the triggering of the $A$-indicator). The experiment investigated the question of which controls will show themselves as characteristic. 

The results are as follows:
\begin{itemize}
\item controls representing very small displacements are characteristic (this is not shown in Fig.\ref{different_controls}).
\item controls that represent very large displacements are not characteristic (shown in Fig.\ref{different_controls}, for example $u = 170$)
\end{itemize}
This reflects the locality of relationships in the structure of the ECG signal: what is close is more predictable than what is far away. Although even what is close is not completely predictable.

The main interest in this experiment, however, is the question, what is the maximum characteristic control? That is, which gaze shift is the largest among those that still lead to predictable results. As can be seen from the Fig.\ref{different_controls}, such a displacement among the displacements to the right is $u = 120$.

It turns out that this shift relative to the R-peak corresponds to the location of the T-wave of this cardiac cycle (T-wave is another stable component of the cardiac cycle along with the R-peak). The search for characteristic control, thus, naturally leads to the modeling of the structure of the cardiac cycle.
\subsection{One-shot learning of indicator B}
Here we demonstrate that learning by one example within the the proposed framework is natural and does not require any modifications.

A control was chosen in which the sample of its sensory results showed a tendency to differ from the general sample of ECG fragments (characteristic control form previous step, $u = 120$ ). In this cloud of 40-dimensional points one point  was randomly chosen. This point, $b$, was used to create a new indicator as described in section \ref{sec::recurr},  i.e. $B(b')$.

One shot learning was organized as follows: in the vicinity of the initial indicator $A$, we searched with a sliding window for such ECG fragments that would trigger that indicator $B(b')$. In this way, we collected a point cloud. After its dimensionality reduction this cloud became dense, shown in Fig.\ref{correction} on right. The center of this cloud was the ideal T-wave of a healthy person, also shown in Fig.\ref{correction}.

\begin{figure*}
\centering
\includegraphics[scale=0.25]{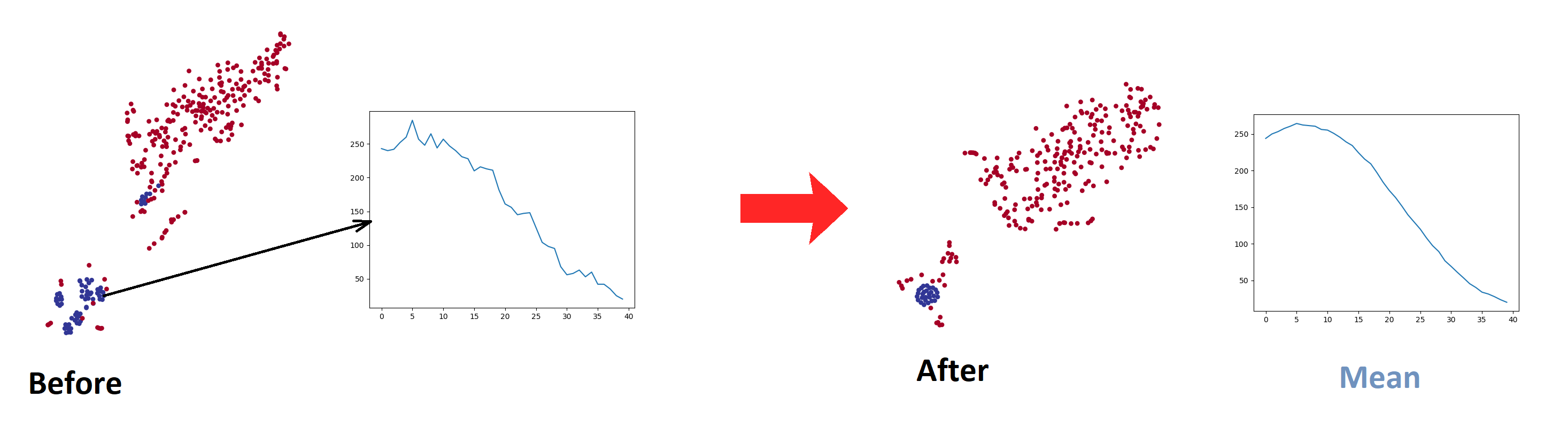}
\caption{\small{An illustration of one shot learning in the proposed framework. In the left it is shown the result of dimensionality reduction  conducted as in fig.\ref{control_dispersion}. The red dots, as usual, represent a sample of random ECG fragments of length 40. The blue ones are chosen according to the control $u = 120$ relative to the R-peak in healthy patients. An arbitrary point from the blue points is selected (shown by a black arrow). It corresponds to the ECG section shown in the figure slightly to the right, and it plays a role of an indicator B. The second "clustering" is obtained in a slightly more complicated way, described in text. Red dots again correspond to random ECG fragments, and blue ones are obtained by cutting out a certain segment of ECG from the vicinity of point B. The rightmost graph shows the center of the new blue cluster. It looks like a part of a perfect T-wave.}}
\label{correction}
\end{figure*}
The experiment demonstrates that one shot active learning in the proposed framework is also natural.

\section{Results and Discussion}
A sketch of adapting the principle of saccadic vision to correct some of the problems inherent in modern deep learning was considered.
It has been shown that the phenomenon of microsaccades occurs naturally in this model, which is consistent with neuroiological data.

The theoretical part describes a learning framework in which the structure of a deep agent grows as new patterns are discovered in the data. The search for new patterns is active - through testing hypotheses about characteristic controls. The growth of the model occurs in the course of an iterative procedure for the formation of new indicators upon detection of characteristic controls for indicators already existing in long-term memory. The long-term memory of the described agent contains information about indicators and controls, thus having a graph-like structure: indicators are nodes, controls are edges, two indicators connected by an edge themselves become a new indicator, that is, a node.

The experimental part shows how the $i+1$-th indicator "grows" from the $i$-th indicator using the example of the ECG. It is shown that if the first indicator is on one of the characteristic waves, which are identified by doctors, then the second also turns out to be on one of the characteristic waves, which has its own name in the medical literature.

If we interpret the proposed recognition process from a Bayesian point of view, then the agent always chooses the hypothesis with the highest likelihood, so that if it is rejected, the Bayesian surprise will be great. An agent built in the framework of the described approach (when controls are explicitly stored in the agent's memory) knows "when it does not know". This is true because the recognition process is a series of hypothesis tests. If each of the hypotheses is true, then the agent's representation can be trusted. If one of the hypotheses was not fulfilled in the experiment, then this fragment of the signal (for which the saccade was intended) contains a surprise, which is an instance of an unknown phenomenon to be studied (i.e., to create a new indicator).

Another possible advantage of the proposed learning ideology is that not the entire model is trained at any given time.  There is no need to set the structure of an agent in advance, unlike many modern networks. The absence of a pre-built network structure opens up the potential for overcoming catastrophic forgetting. This is because it is possible to add new indicators on top of the others into the any place of adaptively growing network structure, since the layered structure no longer exists.

From a philosophical point of view, the advantage of this sketch is that at each step of the agent's recognition of the scene, the result of this recognition represents a natural narrative, just as a path in a semantic graph is a sentence\cite{krishna2017visual} of a natural language. 

The practical results presented in this work are the initial in a series of experiments with ECG and other types of data in the direction of creating learning algorithms in which the agent's data representation contains predictions of sensory control results, and the controls are explicitly stored in the agent's memory.

\bibliographystyle{IEEEtran}
\bibliography{references}

\begin{thebibliography}{10}
\providecommand{\url}[1]{#1}
\csname url@samestyle\endcsname
\providecommand{\newblock}{\relax}
\providecommand{\bibinfo}[2]{#2}
\providecommand{\BIBentrySTDinterwordspacing}{\spaceskip=0pt\relax}
\providecommand{\BIBentryALTinterwordstretchfactor}{4}
\providecommand{\BIBentryALTinterwordspacing}{\spaceskip=\fontdimen2\font plus
\BIBentryALTinterwordstretchfactor\fontdimen3\font minus
  \fontdimen4\font\relax}
\providecommand{\BIBforeignlanguage}[2]{{%
\expandafter\ifx\csname l@#1\endcsname\relax
\typeout{** WARNING: IEEEtran.bst: No hyphenation pattern has been}%
\typeout{** loaded for the language `#1'. Using the pattern for}%
\typeout{** the default language instead.}%
\else
\language=\csname l@#1\endcsname
\fi
#2}}
\providecommand{\BIBdecl}{\relax}
\BIBdecl

\bibitem{lindsay2020convolutional}
G.~W. Lindsay, ``Convolutional neural networks as a model of the visual system:
  past, present, and future,'' \emph{Journal of cognitive neuroscience}, pp.
  1--15, 2020.

\bibitem{karras2019style}
T.~Karras, S.~Laine, and T.~Aila, ``A style-based generator architecture for
  generative adversarial networks,'' in \emph{Proceedings of the IEEE/CVF
  Conference on Computer Vision and Pattern Recognition}, 2019, pp. 4401--4410.

\bibitem{buda2018systematic}
M.~Buda, A.~Maki, and M.~A. Mazurowski, ``A systematic study of the class
  imbalance problem in convolutional neural networks,'' \emph{Neural Networks},
  vol. 106, pp. 249--259, 2018.

\bibitem{kemker2018measuring}
R.~Kemker, M.~McClure, A.~Abitino, T.~Hayes, and C.~Kanan, ``Measuring
  catastrophic forgetting in neural networks,'' in \emph{Proceedings of the
  AAAI Conference on Artificial Intelligence}, vol.~32, no.~1, 2018.

\bibitem{he2021automl}
X.~He, K.~Zhao, and X.~Chu, ``Automl: A survey of the state-of-the-art,''
  \emph{Knowledge-Based Systems}, vol. 212, p. 106622, 2021.

\bibitem{von2015figure}
R.~Von~der Heydt, ``Figure--ground organization and the emergence of
  proto-objects in the visual cortex,'' \emph{Frontiers in psychology}, vol.~6,
  p. 1695, 2015.

\bibitem{krishna2017visual}
R.~Krishna, Y.~Zhu, O.~Groth, J.~Johnson, K.~Hata, J.~Kravitz, S.~Chen,
  Y.~Kalantidis, L.-J. Li, D.~A. Shamma \emph{et~al.}, ``Visual genome:
  Connecting language and vision using crowdsourced dense image annotations,''
  \emph{International journal of computer vision}, vol. 123, no.~1, pp. 32--73,
  2017.

\bibitem{yang2018graph}
J.~Yang, J.~Lu, S.~Lee, D.~Batra, and D.~Parikh, ``Graph r-cnn for scene graph
  generation,'' in \emph{Proceedings of the European conference on computer
  vision (ECCV)}, 2018, pp. 670--685.

\bibitem{xu2017scene}
D.~Xu, Y.~Zhu, C.~B. Choy, and L.~Fei-Fei, ``Scene graph generation by
  iterative message passing,'' in \emph{Proceedings of the IEEE conference on
  computer vision and pattern recognition}, 2017, pp. 5410--5419.

\bibitem{raboh2020differentiable}
M.~Raboh, R.~Herzig, J.~Berant, G.~Chechik, and A.~Globerson, ``Differentiable
  scene graphs,'' in \emph{The IEEE Winter Conference on Applications of
  Computer Vision}, 2020, pp. 1488--1497.

\bibitem{ren2016faster}
S.~Ren, K.~He, R.~Girshick, and J.~Sun, ``Faster r-cnn: towards real-time
  object detection with region proposal networks,'' \emph{IEEE transactions on
  pattern analysis and machine intelligence}, vol.~39, no.~6, pp. 1137--1149,
  2016.

\bibitem{liu2018intriguing}
R.~Liu, J.~Lehman, P.~Molino, F.~P. Such, E.~Frank, A.~Sergeev, and
  J.~Yosinski, ``An intriguing failing of convolutional neural networks and the
  coordconv solution,'' \emph{arXiv preprint arXiv:1807.03247}, 2018.

\bibitem{mnih2014recurrent}
V.~Mnih, N.~Heess, A.~Graves, and K.~Kavukcuoglu, ``Recurrent models of visual
  attention,'' \emph{arXiv preprint arXiv:1406.6247}, 2014.

\bibitem{vaswani2017attention}
A.~Vaswani, N.~Shazeer, N.~Parmar, J.~Uszkoreit, L.~Jones, A.~N. Gomez,
  L.~Kaiser, and I.~Polosukhin, ``Attention is all you need,'' \emph{arXiv
  preprint arXiv:1706.03762}, 2017.

\bibitem{fabius2020low}
J.~H. Fabius, A.~Fracasso, D.~J. Acunzo, S.~Van~der Stigchel, and D.~Melcher,
  ``Low-level visual information is maintained across saccades, allowing for a
  postsaccadic handoff between visual areas,'' \emph{Journal of Neuroscience},
  vol.~40, no.~49, pp. 9476--9486, 2020.

\bibitem{yarbus2013eye}
A.~L. Yarbus, \emph{Eye movements and vision}.\hskip 1em plus 0.5em minus
  0.4em\relax Springer, 2013.

\bibitem{tian2018dynamics}
X.~Tian, M.~Yoshida, and Z.~M. Hafed, ``Dynamics of fixational eye position and
  microsaccades during spatial cueing: the case of express microsaccades,''
  \emph{Journal of neurophysiology}, vol. 119, no.~5, pp. 1962--1980, 2018.

\bibitem{friston2012perceptions}
K.~Friston, R.~Adams, L.~Perrinet, and M.~Breakspear, ``Perceptions as
  hypotheses: saccades as experiments,'' \emph{Frontiers in psychology},
  vol.~3, p. 151, 2012.

\bibitem{moser2008place}
E.~I. Moser, E.~Kropff, and M.-B. Moser, ``Place cells, grid cells, and the
  brain's spatial representation system,'' \emph{Annu. Rev. Neurosci.},
  vol.~31, pp. 69--89, 2008.

\bibitem{long2018novel}
X.~Long and S.-J. Zhang, ``A novel somatosensory spatial navigation system
  outside the hippocampal formation,'' \emph{bioRxiv}, p. 473090, 2018.

\bibitem{murata1992effects}
K.~Murata, P.~J. Landrigan, and S.~Araki, ``Effects of age, heart rate, gender,
  tobacco and alcohol ingestion on rr interval variability in human ecg,''
  \emph{Journal of the autonomic nervous system}, vol.~37, no.~3, pp. 199--206,
  1992.

\bibitem{malali2020supervised}
A.~Malali, S.~Hiriyannaiah, G.~Siddesh, K.~Srinivasa, and N.~Sanjay,
  ``Supervised ecg wave segmentation using convolutional lstm,'' \emph{ICT
  Express}, vol.~6, no.~3, pp. 166--169, 2020.

\bibitem{acharya2017application}
U.~R. Acharya, H.~Fujita, S.~L. Oh, Y.~Hagiwara, J.~H. Tan, and M.~Adam,
  ``Application of deep convolutional neural network for automated detection of
  myocardial infarction using ecg signals,'' \emph{Information Sciences}, vol.
  415, pp. 190--198, 2017.

\bibitem{yildirim2018efficient}
O.~Yildirim, R.~San~Tan, and U.~R. Acharya, ``An efficient compression of ecg
  signals using deep convolutional autoencoders,'' \emph{Cognitive Systems
  Research}, vol.~52, pp. 198--211, 2018.

\bibitem{sereda2020problems}
I.~Sereda, S.~Alekseev, A.~Koneva, A.~Khorkin, and G.~Osipov, ``Problems of
  representation of electrocardiograms in convolutional neural networks,'' in
  \emph{2020 International Joint Conference on Neural Networks (IJCNN)}.\hskip
  1em plus 0.5em minus 0.4em\relax IEEE, 2020, pp. 1--6.

\bibitem{hafting2005microstructure}
T.~Hafting, M.~Fyhn, S.~Molden, M.-B. Moser, and E.~I. Moser, ``Microstructure
  of a spatial map in the entorhinal cortex,'' \emph{Nature}, vol. 436, no.
  7052, pp. 801--806, 2005.

\bibitem{zhang2019making}
R.~Zhang, ``Making convolutional networks shift-invariant again,'' in
  \emph{International Conference on Machine Learning}.\hskip 1em plus 0.5em
  minus 0.4em\relax PMLR, 2019, pp. 7324--7334.

\bibitem{kalyakulina2020ludb}
A.~I. Kalyakulina, I.~I. Yusipov, V.~A. Moskalenko, A.~V. Nikolskiy, K.~A.
  Kosonogov, G.~V. Osipov, N.~Y. Zolotykh, and M.~V. Ivanchenko, ``Ludb: a new
  open-access validation tool for electrocardiogram delineation algorithms,''
  \emph{IEEE Access}, vol.~8, pp. 186\,181--186\,190, 2020.

\end{thebibliography}

\end{document}